\documentclass{article}
\usepackage{ijcai21}

\usepackage{times}

\usepackage{soul}
\usepackage{url}
\usepackage[hidelinks]{hyperref}
\usepackage[utf8]{inputenc}
\usepackage[small]{caption}
\usepackage{graphicx}
\usepackage{amsmath}
\usepackage{booktabs}
\usepackage[switch]{lineno}

\usepackage{bm}
\usepackage{algorithm}
\usepackage{algpseudocode}
\usepackage{multirow}
\usepackage{amssymb}

\title{Neighbourhood-guided Feature Reconstruction for Occluded Person Re-Identification }
\author {
 Anonymous
 \\
}

\author{
Shijie Yu$^{1,2}$\footnote{sj.yu@siat.ac.cn}\and
Dapeng Chen$^3$\and
Rui Zhao$^{3}$\and
Haobin Chen$^{1,2}$\And
Yu Qiao$^{1,4}$\footnote{Corresponding Author}
\affiliations
$^1$ShenZhen Key Lab of Computer Vision and Pattern Recognition, SIAT-SenseTime Joint Lab, Shenzhen Institute of Advanced Technology, Chinese Academy of Sciences\\
$^2$University of Chinese Academy of Sciences, China\\
$^3$SenseTime Group Limited\\
$^4$Shanghai AI Lab, Shanghai, China\\
\emails

}

\begin{document}
\maketitle

\begin{abstract}
Person images captured by surveillance cameras are often occluded by various obstacles, which lead to defective feature representation and harm person re-identification (Re-ID) performance. To tackle this challenge, we propose to reconstruct the feature representation of occluded parts by fully exploiting the information of its neighborhood in a gallery image set. Specifically, we first introduce a visible part-based feature by body mask for each person image. Then we identify its neighboring samples using the visible features and reconstruct the representation of the full body by an outlier-removable graph neural network with all the neighboring samples as input.  Extensive experiments
show that the proposed approach obtains significant improvements.  In the large-scale Occluded-DukeMTMC benchmark, our approach achieves 64.2\% mAP and 67.6\% rank-1 accuracy which outperforms the state-of-the-art approaches by large margins, \emph{i.e}.,20.4\% and 12.5\%, respectively, indicating the effectiveness of our method on occluded Re-ID problem.
\end{abstract}

\section{Introduction}

\begin{figure}
    \centering
    \includegraphics[width=0.95\linewidth]{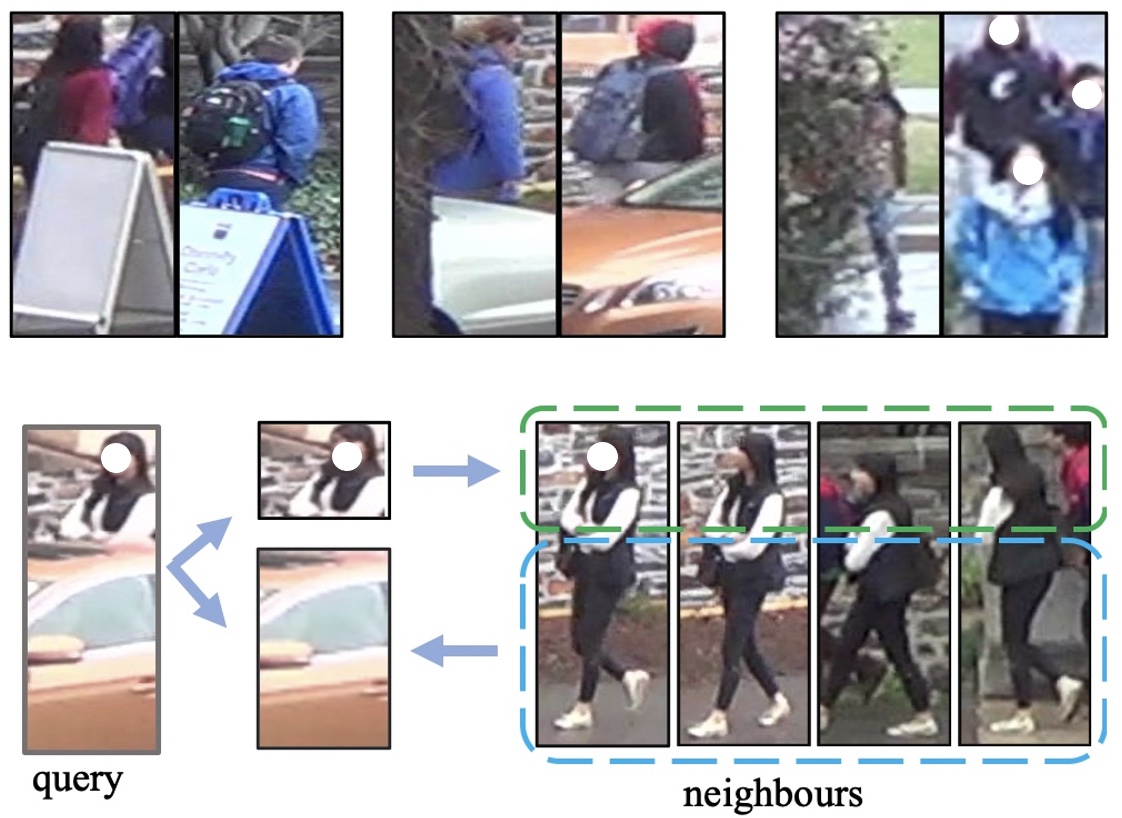}
    \caption{\textbf{Top:} Several examples of occluded person images are captured in real-world surveillance scenarios. \textbf{Bottom:} The illustration of our motivation in this paper.  The occluded query image is inferred to obtain the occluded and non-occluded parts.  We employ the person of interest part to obtain its neighborhoods (the green box) from a gallery set, and the occlusion's counterparts (the blue box) in neighborhoods are used to reconstruct the missing body. }
    \label{fig:my_label}
    \vspace{-1.0em}
\end{figure}

Person re-identification aims to accurately associate the corresponding person's images across non-overlapped cameras and has many applications in public security, automated retailers, and smart city.  Although it has achieved significant progress in general matching accuracy, the current methods are assuming that  the person images are visible without any occlusions. In practice,  person images from surveillance cameras are often occluded by other persons, vehicles, or even plants along the road in real-world scenarios as shown in the top of Fig. \ref{fig:my_label}, which brings obstacles to correctly match with another image of the same person. Hence, it is urgent to develop effective methods  to represent and match with the occluded images.

Two critical problems need to be addressed in occluded person Re-ID. Firstly, the Re-ID model should be able to distinguish the visible body regions and obstacles. Secondly, the Re-ID models should be able to match person samples with different occluded parts and viewpoints. Previous methods \cite{Miao_2019_ICCV,Wang_2020_CVPR,Gao_2020_CVPR,Zhuo2018OccludedPR,He_2019_ICCV,Huang_2018_CVPR} design various strategies to avoid matching with occluded parts.  However, as the locations and sizes of occlusion parts vary among different images, it lacks a unified criterion to reflect the confidence of two images belonging to the same person.


In this paper, we propose to reconstruct the representation of the whole body for an occluded person by exploiting the information of its neighbors. By assuming there exist non-occluded images are belonging to the same person for each occluded image,  we aim to first find the non-occluded parts of the same person from other images, then utilize the part features to reconstruct the part features of the occluded images.  In this way, all the images have equally comparable features, which will bridge the matching gap between occluded and non-occluded images. More specifically, we employ a part-based model named feature extractor network (FEN), which is similar to MGN \cite{Wang2018LearningDF} and PCB \cite{Sun_2018_ECCV}, to better learn the part-based representations of person images. Besides, we learn an occlusion estimator named occlusion-aware network (OAN) to determine whether each part is occluded or not. In particular, the occlusion estimator takes the human masks generated by the Mask-RCNN \cite{Mask_RCNN} as input. The neighborhood of an occluded image is collected according to the occlusion states, which is the intersection of neighbors of non-occluded parts and is used to reconstruct the full-body feature representation of the image.

An outlier-removable graph neural network is therefore proposed for the feature reconstruction. It consists of an outlier-aware module and a feature transformation module. The outlier-aware module allocates a confidence score for each sample in the neighbourhood, indicating whether a node belonging to the same person or not. The feature transformation module updates the nodes and edges of the selected neighbours to obtain the feature representation of the interested image. The two modules work alternatively.  All sub-features of an image are reconstructed by the corresponding sub-features of highlighted neighbours. We concatenate all the reconstructed sub-features as the image's final representation for retrieval. Extensive experiments on three occluded person benchmarks have shown that the approach is able to handle the occluded person Re-ID problem effectively. We have achieved 64.2\% mAP and 67.6\% Rank-1 accuracy on Occluded-DukeMTMC, which outperforms the HONet \cite{Wang_2020_CVPR} of the state-of-the-arts more than 20.4\% and 12.5\%, respectively. In summary, the main contributions are two-folder: 
\begin{itemize}
    \item We present a new perspective that the representation of an occluded person image can be reconstructed by the non-occluded parts of its neighbours.  
    \item We propose an outlier-removable graph neural network to automatically highlight the neighbours belonging to the same person, and transform their features to approximate non-occluded representation of an occluded person image. 
    \item The proposed method achieves significant improvements on Occluded-DukeMTMC which is a large-scale occluded Re-ID benchmark. Meanwhile, on small-scale occluded Re-ID and traditional Re-ID benchmarks, it also enables to obtain competitive results.
\end{itemize}


\section{Related Work}
\noindent \textbf{Occluded Person Re-ID} 
The problem of the occluded person Re-ID has recently attracted more and more research interests. A straightforward solution is to learn a powerful feature representation that can bridge the gap between the occluded and non-occluded person images. For this purpose, \cite{Zhuo2018OccludedPR} attempted to simulate the occlusion and used multi-task losses (\emph{i.e}, occluded/non-occluded binary classification (OBC) loss and softmax loss to train the model. Pose estimation and human parsing are utilized as the extra information to distinguish the occlusions. e.g., \cite{He_2019_ICCV}  used human parsing information to guide the network to be aware of the target persons in occluded images. \cite{Gao_2020_CVPR} proposed to involve the human key-points to align the local features when matching two images. Our approach is most relevant to \cite{Wang_2020_CVPR}, which builds a graph within an image for feature learning and matching with human topology.  Different from their approach, we build a graph among neighboring images, using the part-features of other images to reconstruct the feature of the occluded parts.


\noindent \textbf{Part-based Person Re-ID}
Part-based person re-id approaches \cite{Sun_2018_ECCV,Wang2018LearningDF,Fu2019HorizontalPM,Zhao_2017_ICCV,Part_1,part_2,part_3}, exploit local descriptors from different regions to enhance the discriminative ability of the feature representation. One simple way to generated parts is to split the person images or feature maps into multiple stripes.  \cite{Sun_2018_ECCV} split the feature maps into $p$ parts horizontally,  where each part reserves its classifier. \cite{Wang2018LearningDF} designed a network with three branches and each branch exploits different granularity local features. The part features can also be extracted by pose driven RoI, human parsing results or learning attention regions. \cite{part_2} proposed to generate part maps from prior pose information and then aggregate all parts with a bilinear pooling. \cite{Kalayeh_2018_CVPR} utilized human semantic parsing results to extract body part features. We also adopt the stripe-based strategy, but utilize only local sub-features of MGN \cite{Wang2018LearningDF}. An occlusion estimator is additionally learned to indicate the occlusion state of each part.  

\begin{figure*}[tbp]
    \centering
    \includegraphics[width=0.95\linewidth]{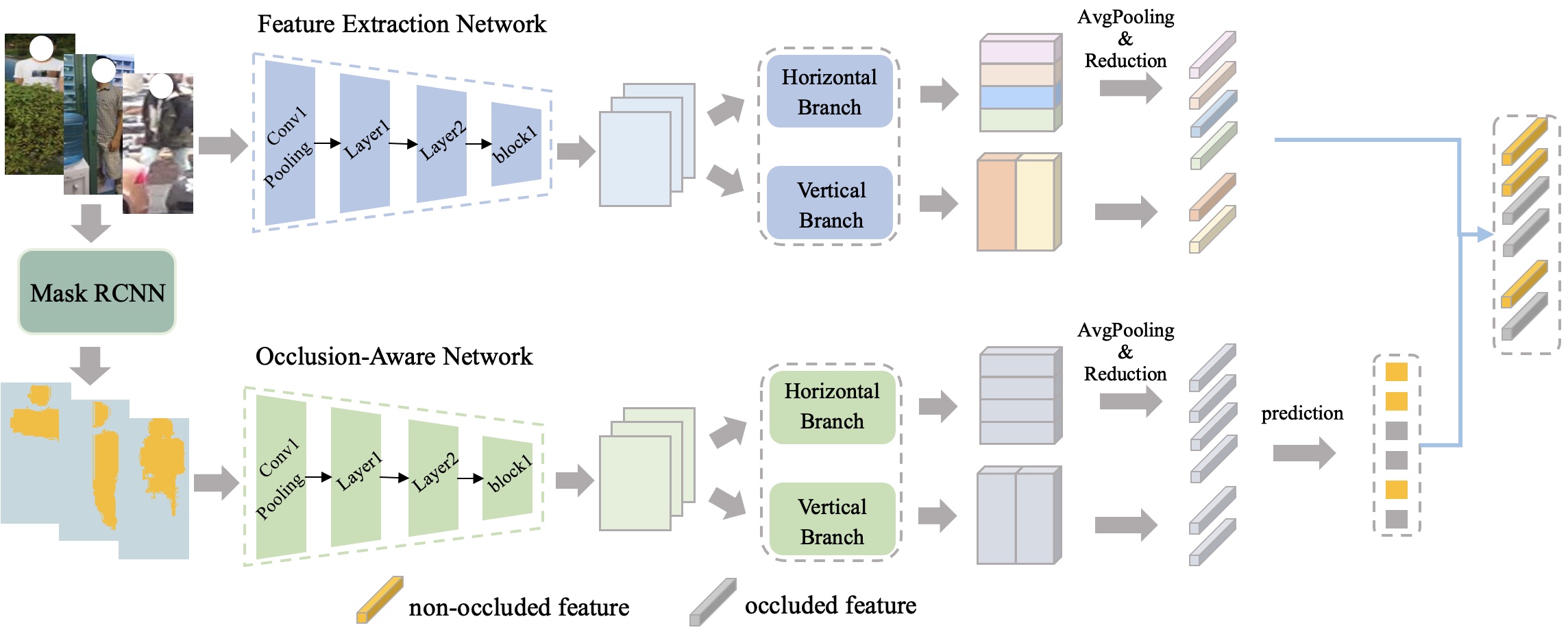}
    \vspace{-0.5em}
    \caption{The architectures of feature extraction network (FEN) and occlusion-aware network (OAN). Both of FEN and OAN take ResNet-50 as the backbones and are split into two branches after the 1st block of the 3rd layer. FEN takes person images as input to obtains part representations for each image, while OAN takes human masks as input to estimate whether the corresponding parts are occluded or not.}
    \label{fig:network}
    \vspace{-1.0em}
\end{figure*}


\section{Methodology}

We propose neighborhood-guided feature reconstruction to tackle the challenges in the occluded person Re-ID. Our method includes two stages. In the first stage,  we learn a part-based representation as well as the corresponding occlusion states of an interested image.  In the second stage, we find high-quality neighbors of the image by using the features of non-occluded parts. An outlier-removable network is further developed to highlight the neighbors belonging to the same person, then encode the new features with the neighbors.

\subsection{Occlusion-Aware Part-based Representation}

\subsubsection{Part-based Representation}
Let $I$ denote an image, we feed the image into a feature extraction network (FEN) to obtain $M$ sub-features $\{ \mathbf{x}_{i}(I)\}_{i=1}^{M}$ where $\mathbf{x}_{i}(I)$ indicates the $i$th sub-feature.  
More specifically, the FEN is inspired by the Multiple Granularity Network (MGN) \cite{Wang2018LearningDF}, which is one of the most effective part-based models. The backbone of FEN is modified from ResNet-50 \cite{ResNet}. After the 1st block of 3rd layer, we separate the network into two branches. To better learn part features, the two branches focus on different regions and do not share the weights. The first branch is used to extract the horizontal sub-features, and its feature map is split into four parts horizontally. The second branch is used to handle the case when a person is vertically occluded, and its feature map is vertically split into two parts to obtain the vertical sub-features.  We therefore obtain 6 sub-features in total, including 4 horizontal sub-features and 2 vertical sub-features, for each image via FEN.

To learn FEN, we employ both identification loss and triplet loss. The identification loss is a general cross-entropy loss function, and we define $\mathcal{L}_{id}(\mathbf{x}_{i}(I_{a}))$ for the $i$th sub-feature of the image $I_{a}$. The triplet loss $\mathcal{L}_{tri}( \mathbf{x}_{i}(I_{a}))$ applied on the $i$th sub-feature, is defined as: $\max\{\psi( \mathbf{x}_{i}(I_{a}),  \mathbf{x}_{i}(I_{p}))-\psi( \mathbf{x}_{i}(I_{a}),  \mathbf{x}_{i}(I_{n})) + \eta, 0\}$. In greater details,  $\psi(\cdot)$ is to compute the distance of pairs. $\mathcal{I}_p$ denotes farthest positive sample while $\mathcal{I}_n$ is the nearest negative sample w.r.t. the image $\mathcal{I}_a$. The hyper-parameters $\eta$ is the margin set to be 0.3. The joint loss function for FEN is defined as follows:
\begin{equation}
\begin{split}
    \mathcal{L}_{FEN}(\mathcal{I}_{a}) &= \frac{1}{M}\sum_{i=1}^{M}\mathcal{L}_{id}(\mathbf{x}_{i}(I_{a}))+\mathcal{L}_{tri}( \mathbf{x}_{i}(I_{a})).
\end{split}
\end{equation}
\subsubsection{Occlusion Estimation} To determine whether each part is occluded or not, we propose occlusion-Aware Network (OAN) associated with the FEN. The OAN has a similar network architecture with the FEN as shown in Fig. \ref{fig:network}. Note that, the OAN takes human mask images generated by Mask-RCNN as inputs. The OAN produces a vector $y(\mathcal{I}_{a})$ for the image $\mathcal{I}_{a}$ to show the occluded state of each part. A binary loss function is designed for the training of OAN is therefore given by:
\begin{equation}
\begin{split}
    \mathcal{L}_{OAN}(\mathcal{I}_{a}) &= \hat{y}(\mathcal{I}_{a}) \log y(\mathcal{I}_{a}) \\
    &+(1-\hat{y}(\mathcal{I}_{a})) \log(1 - y(\mathcal{I}_{a})). 
    \end{split}
\end{equation}
With the predicted $y(\mathcal{I}_{a})$, we obtain the binary occlusion state by filtering with a threshold $0.5$, and the binary occlusion vector is denoted by $m(\mathcal{I}_{a})$. 
\begin{figure*}
    \centering
    \includegraphics[width=0.95\linewidth]{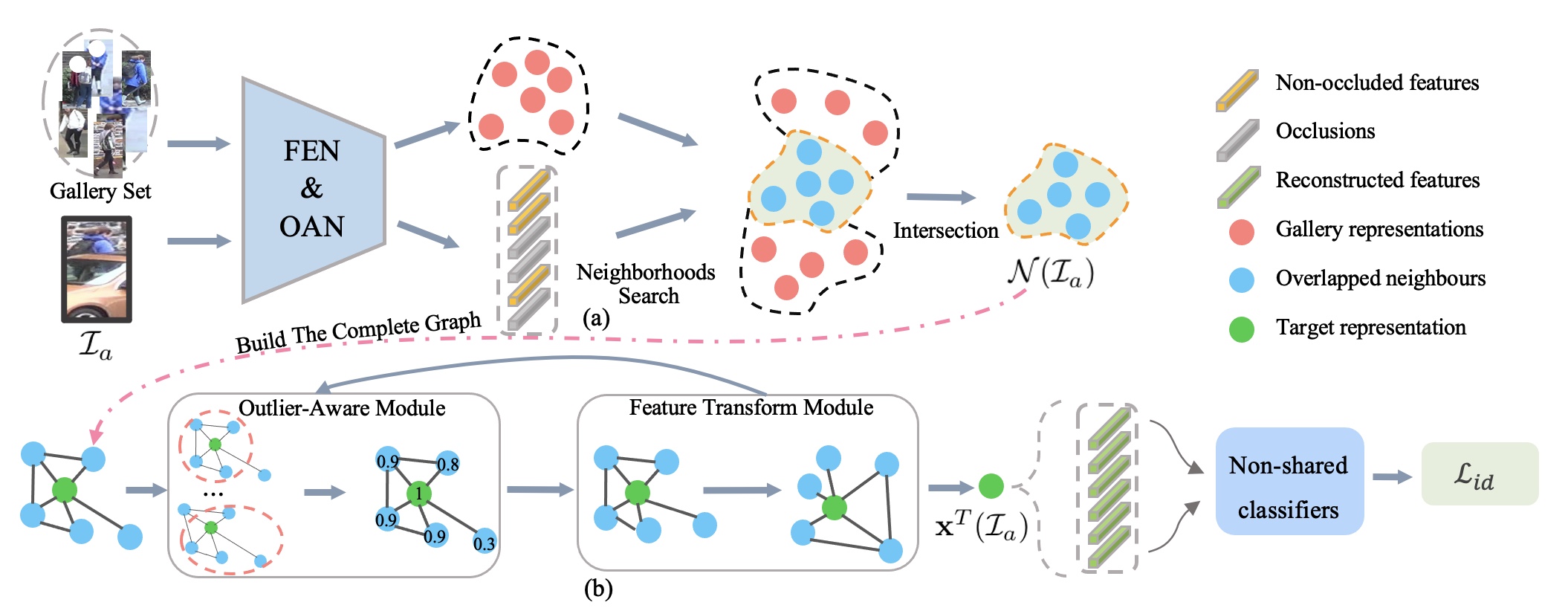} \vspace{-1em}
    \caption{The overall framework of the proposed method. Given an occluded image and a galley set, (a) shows the procedure of using FEN and OAN to obtain the occlusion-aware representation as well as the generation of the non-occluded sub-features. (b) is the architecture of OR-GNN. We first build the graph by the neighborhood and the target (the green dot) which is initialized by neighbours' average representation. In the following, the outlier-aware module is to obtain the confidence for each node and the feature transform module is to aggregate neighborhood. Finally, OR-GNN generates the reconstruction representation of the input occluded person image.}
    \label{fig:rgnn}
    \vspace{-0.5em}
\end{figure*}

\subsection{Feature Reconstruction via OR-GNN}
With the FEN and the OAN, we obtain sub-features as well as the occlusion states of all parts. According to these information, neighbourhood search is applied to find more non-occluded feature from a same person. An outlier-removable graph neural network (OR-GNN) is therefore introduced to both highlight more reliable neighbours and reconstruct the features of the interested image. As the reconstructed feature overcomes the invisibility of the occluded image, the new features can be used for general person matching.  The overall framework of our proposed method is shown in Fig. \ref{fig:rgnn}.


\subsubsection{Neighborhood Search} We utilize the sub-features of the non-occluded part to search the neighborhood from the gallery set. Taking image $\mathcal{I}_{a}$ for example, the neighbourhood of a non-occluded part $k$ is the $K$ nearest neighbours searched by the sub-feature of the same part under the control of a similarity threshold $\theta$, and the neighborhood is denoted by $\mathcal{N}(\mathbf{x}_{k}(\mathcal{I}_{a}))$. Because each image may have multiple non-occluded parts, it will generate multiple neighbourhood of sub-features. We select the intersection of these sets as the neighbourhood of the image $\mathcal{I}_{a}$, 
\begin{equation}
\mathcal{N}(\mathcal{I}_{a}) =  \bigcap_{m_{k}(\mathcal{I}_{a}) \neq 0} \mathcal{N}(\mathbf{x}_{k}(\mathcal{I}_{a})), 
\end{equation}
where $m_{k}(\mathcal{I}_{a})$ indicates the occlusion state of the $k$th part.



\subsubsection{Representation Reconstruction}  Given an image $\mathcal{I}_{a}$ and its neighbourhood $\mathcal{N}(\mathcal{I}_{a})$ 
as the input, we can build multiple complete graphs with different sub-features. Each graph is fed into the \textbf{O}utlier \textbf{R}emovable \textbf{G}raph \textbf{N}eural \textbf{N}etwork (OR-GNN), which can both remove the outliers and aggregate the sub- features with the information of its neighbourhood. Each layer of OR-GNN consists of an outlier-aware module and a feature transformation module. Alg. \ref{alg:or-gnn} shows the whole pipeline of the T-layer OR-GNN.

The outlier-aware module is to estimate the confidence of each node belonging to the same person with $\mathcal{I}_{a}$. Consider the $k$th sub-features at the $t$th layer of the GNN, \emph{i.e.}, the input of the OR-GNN is represented as $\mathcal{D}^{t}_{k}=\{\mathbf{x}^{t}_{k}(\mathcal{I})| \mathcal{I}\in \mathcal{N}(\mathcal{I}_{a}) \cup {\mathcal{I}_a}\}$, where $\mathbf{x}_k(\mathcal{I}_a)$ is initialized by $ \sum_{\mathcal{I}\in \mathcal{N}(\mathcal{I}_{a})}\mathbf{x}_k^1(\mathcal{I})/|\mathcal{N}(\mathcal{I}_a)|$,  we obtain the confidence of $\mathbf{x}^{t}_{k}(\mathcal{I})$ by computing the cosine similarity between its feature and all other features:
\begin{equation}
   c^{t}_{k}(\mathcal{I}) = \frac{\mathbf{x}^{t}_{k}(\mathcal{I}) \cdot \rho(\mathcal{D}^{t}_{k}\setminus \mathbf{x}^{t}_{k}(\mathcal{I}))}{\|\mathbf{x}^{t}_{k}(\mathcal{I}) \|\cdot\| \rho(\mathcal{D}^{t}_{k}\setminus \mathbf{x}^{t}_{k}(\mathcal{I})) \|}, 
   \label{eq:confidence}
\end{equation}
where $\rho(\cdot)$ is used to compute the average feature and $\mathcal{D}^{t}_{k}\setminus \mathbf{x}^{t}_{k}(\mathcal{I})$ denotes the set where $\mathbf{x}^{t}_{k}(\mathcal{I})$ is excluded from $\mathcal{D}^{t}_{k}$. We assume the distribution of the outliers is different from the features belonging to the same person. The outliers will have low similarity with most of the other samples. If the confidence $c^{t}_{k}(\mathcal{I})$ is low,  the relevance between $\mathbf{x}^{t}_{k}(\mathcal{I})$ and the other samples in the neighbourhood is less, thus it is more likely to be the outliers within the neighbourhood.

The confidence is involved into the feature transformation module. The transformed feature is obtained by both feature aggregation and linear transformation:
\begin{equation}\small
\mathbf{x}^{t+1}_{k}(\mathcal{I}_{i}) \!=\! \phi( \mathbf{W}^{t}_{k}\frac{\sum_{\mathcal{I}_{j} \in \mathcal{N}(\mathcal{I}_{i})} c_{k}^{t}(\mathcal{I}_{j}) a_{k}^{t}(\mathcal{I}_{i}, \mathcal{I}_{j})\mathbf{x}_{k}^{t}(\mathcal{I}_{j})}{
\sum_{\mathcal{I}_{j} \in \mathcal{N}(\mathcal{I}_{i})} c_{k}^{t}(\mathcal{I}_{j}) a_{k}^{t}(\mathcal{I}_{i}, \mathcal{I}_{j}) 
}),
 \label{eq:nodes}
\end{equation}
where $\mathbf{W}^{t}_{k}$ is the parameter for learning transformation and $\phi(\cdot)$ is the ReLU function. $a_{k}^{t}(\mathcal{I}_{i}, \mathcal{I}_{j})$ is the affinity for the feature aggregation. The affinity $a_{k}^{t}(\mathcal{I}_{i}, \mathcal{I}_{j})$ is computed by the features $x_{k}^{t}(\mathcal{I}_{i})$ and $x_{k}^{t}(\mathcal{I}_{j})$:
\begin{equation}
  \begin{split}
      d_{k}^{t}(\mathcal{I}_{i}, \mathcal{I}_{j}) &= x_{k}^{t}(\mathcal{I}_{i}) -  x_{k}^{t}(\mathcal{I}_{j})\\
      a_{k}^{t}(\mathcal{I}_{i}, \mathcal{I}_{j}) &= \sigma(V_{k}^{t}(  d_{k}^{t}(\mathcal{I}_{i}, \mathcal{I}_{j}) \circ d_{k}^{t}(\mathcal{I}_{i}, \mathcal{I}_{j}))+ b_{k}^{t}), 
  \end{split}
  \label{eq:edges}
\end{equation}
where $V^{t}_{k}$ and $b^{t}_{k}$ are parameters of linear transformation to obtain pairwise affinity. With features $x_{k}^{t}(\mathcal{I}_{i})$ and $x_{k}^{t}(\mathcal{I}_{j})$, we perform element-wise square of their distance, project the obtained vector to a scalar value, and normalize the scalar value to be within $(0, 1)$ with the sigmoid function $\sigma(\cdot)$. We repeat the outlier-aware module and the feature transformation module for $T$ times, and only take the feature embedding of the interested image, \emph{i.e.}, $\mathbf{x}_{k}^{T}(\mathcal{I}_{a})$, as the reconstructed sub-feature for the image $\mathcal{I}_{a}$.

\begin{algorithm}[tbp]
    \caption{The process of OR-GNN for inference}
    \label{alg:or-gnn}
    \begin{algorithmic}
        \Require nodes of $k$th sub-graph $\mathcal{D}_{k}^0=\{\mathbf{x}_{k}^0(\mathcal{I})| \mathcal{I}\in \mathcal{N}(\mathcal{I}_{a}) \cup \{\mathcal{I}_{a}\} \}$
        \Ensure The reconstruction representation $\mathbf{x}^T_k(\mathcal{I}_a)$
        \For{$i = 1, 2, ...,T $}
        \State 1. Compute the nodes' confidence w.r.t. Eq. \ref{eq:confidence}.
        \State 2. Obtain pairwise affinity by Eq. \ref{eq:edges}.
        \State 3. Perform feature transformation w.r.t Eq. \ref{eq:nodes}.
        \EndFor
    \end{algorithmic}
\end{algorithm}

 Different kinds of sub-features are fed to different OR-GNN. We learn the multiple OR-GNNs simultaneously, and employ the cross-entropy loss function:
 \begin{equation}
     \mathcal{L}_{OR-GNN}(\mathcal{I}_{a}) = \frac{1}{M}{\sum_{k=1}^{M} \mathcal{L}_{id}(\mathbf{x}_{k}^{T}(\mathcal{I}_{a}))},
 \end{equation}
 where $M$ is the number of the sub-features. 
 
 
 \section{Experiments}

In this section, we demonstrate the effectiveness of the proposed method for occluded person Re-ID problem by comparing with the state-of-the-art methods as well as conducting extensive ablation studies.



 
\subsection{Datasets \& Evaluation Protocol}
\textbf{Occluded-DukeMTMC} is built from DukeMTMC \cite{Zheng_2017_ICCV} and contains 15,618 training images of 708 persons, 2,210 query images of 519 persons and 17,661 gallery images of 1110 persons. Notice that, person images in the query set are all occluded by various obstacles. Besides, compared with other occluded datasets, Occluded-DukeMTMC is larger, and closer to the real-world surveillance scenarios. 

\noindent \textbf{Occluded-REID} contains 200 persons and 2000 images and each identity includes 10 images, 5 full-body images and 5 occluded images with different occlusions, respectively. Usually, the occluded images are considered as the query set while the all full-body images are put into gallery set.

\noindent \textbf{Partial-REID} contains 60 persons and 600 images which is collected from the campus of a university. Similarly, each identity in Partial-REID has 5 full-body images in the gallery set and 5 occluded images in the query set. In particular, unlike \cite{DSR}, we use the occluded images rather than the partial images with occluded parts manually cropped. 

\noindent \textbf{Market1501} contains 32,668 images of 1,501 identities captured from 6 cameras. It is split into two parts: 12,936 images of 751 identities for training and 19,732 images of 750 identities for testing. In testing, 3,368 images are used as query set and the remaining images are gallery images.

\noindent \textbf{DukeMTMC} contains 36,411 images of 1,812 identities in total. In greater details, 16,522 images of 702 identities are used for training, 2,228 images of another 702 identities are used as query images, and the remaining 17,661 images are gallery images.

\noindent\textbf{Evaluation Protocol} We employ the widely used evaluation metrics, including Cumulative Matching Characteristic (CMC) curves and mean average precision (mAP), to evaluate the proposed method with different settings.

\begin{table}[t]
 \begin{center}
  \begin{tabular}
    {l|c |c| c| c}
    \hline
     \textbf{Method} & mAP & Rank-1 & Rank-5 & Rank-10  \\ 
    \hline
    \hline
    LOMO+XQDA&5.0&8.1&17.0&22.0\\
    DSR &30.4&40.8&58.2&65.2 \\
    SFR &32.0&42.3&60.3&67.3 \\
    HACNN &26.0&34.4&51.9&59.4 \\
    Adver Occluded&32.2&44.5&-&- \\
    PCB&33.7&42.6&57.1&62.9 \\
    PGFA&37.3&51.4&68.6&74.9 \\
    HONet&43.8&55.1&-&- \\
    \hline
    \hline
    Baseline &42.3&48.7&65.9&71.7 \\
    Ours&\textbf{64.2}&\textbf{67.6}&\textbf{77.3}&\textbf{79.5} \\
    Ours (UB)&67.9&73.5&81.2&82.4\\
    \hline
 \end{tabular}
 \caption{Performance comparison with the state-of-the-arts on large-scale benchmark of Occluded-DukeMTMC. The SOTA methods include LOMO+XQDA \protect\cite{LOMO}, DSR \protect\cite{DSR}, SFR \protect\cite{SFR},HACNN \protect\cite{HACNN}, Adver Occluded \protect\cite{Huang_2018_CVPR}, PCB \protect\cite{Sun_2018_ECCV}, PGFA \protect\cite{Miao_2019_ICCV}, and HONet \protect\cite{Wang_2020_CVPR}. 
 ``UB" denotes ``upper bound" which means we employ the ground truth label to remove the outliers of neighbours.}
 \label{tab:occluded-duke} \vspace{-1em}
 \end{center} 
\end{table}

\begin{table}[t]
 \begin{center}
  \begin{tabular}
    {l|c |c| c| c}
    \hline
    \multicolumn{1}{l|}{\multirow{2}{*}{\textbf{Method}}}&\multicolumn{2}{c|}{\textbf{Partial-REID}}&\multicolumn{2}{c}{\textbf{Occluded-REID}}  \\   
    \cline{2-5}
    &mAP&Rank-1&mAP&Rank-1 \\ 
    \hline
    \hline
    AMC+SWM&-&36.0&27.3&31.1\\
    PCB&54.7&56.3&38.9&41.3 \\
    DSR &-&43.0&62.8&\textbf{72.8} \\
    SFR &-&56.9&60.3&67.3 \\
    PGFA&-&68.0&-&- \\
    PVPM&72.3&\textbf{78.3}&61.2&70.4 \\
    \hline
    \hline
    Baseline &62.6&69.2&58.3&65.1 \\
    Ours&\textbf{74.7}&75.8&\textbf{67.3}&68.8 \\
    \hline
 \end{tabular}
 \caption{Results on small-scale datasets Partial-REID and Occluded-REID. We compare with AMC+SWM \protect\cite{Zheng_2015_ICCV}, PCB \protect\cite{Sun_2018_ECCV}, DSR \protect\cite{DSR}, SFR \protect\cite{SFR}, PGFA \protect\cite{Miao_2019_ICCV} and PVPM \protect\cite{Gao_2020_CVPR}.}
 \label{tab:small_dataset} \vspace{-1em}
 \end{center} 
\end{table}

\subsection{Implementation Details}
We employ Adam to optimize the parameters of FEN and OAN, and it lasts 120 epochs totally. The learning rate is set to 3.5e-4 initially and decays to 0.1 of the previous at 40 and 70 epoch. 1 GTX-1080TI GPUs are used and the batch size of each GPU is set to 64. In particular, each batch includes 8 persons and each person has 4 images. The input images are resized to  $384 \times 128$ with data augmentations, such as random erasing and random flipping. 

For the training of OR-GNN, we obtain neighborhoods with $K\!=\!30$ and $\theta\!=\!0.7$. The OR-GNN contains 2 layers. Also, we use Adam to optimize its parameters which lasts 120 epochs. The learning rate is initialized as 3.5e-4 and decays to 0.1 of the previous at 40 and 70 epochs. Moreover, 1 GTX-1080TI GPU is used to train OR-GNN and a batch of training data includes 16 persons and each person has 4 neighbours sets. In the phrase of inference, we set $K\!=\!10, \theta\!=\!0.7$ for Occluded-DukeMTMC, $K\!=\!4, \theta\!=\!0.8$ for Occluded-REID,  Partial-REID, Market1501 and DukeMTMC.

To better demonstrate the effectiveness of the proposed method, we conduct baseline method for analysis. \textbf{Baseline} indicates retrieving person images directly by concatenating all sub-features of FEN.





\subsection{Comparison with the State-of-the-arts}
\noindent \textbf{Results on Occluded-DukeMTMC} We first evaluate the proposed approach on Occluded-DukeMTMC, where each person has about 15 images in the gallery. Tab.\ref{tab:occluded-duke} shows the comparison with several state-of-the-art (SOTA) methods. Obviously, OR-GNN achieves a superior performance of 64.2 \% mAP and 67.6\% Rank-1 accuracy, which outperforms the baseline (\emph{i.e.}, the FEN only) by 21.9\% and 18.9\% in terms of mAP and Rank-1 accuracy. Compared with current SOTA method of HONet \cite{Wang_2020_CVPR}, our approach achieves 20.4\% mAP and 12.5\% Rank-1 accuracy gains respectively. The significant improvement demonstrates that the occlusion-aware part-based representation is much effective to address occluded person Re-ID problem and the new representation reconstructed by neighborhood of an occluded image is more discriminative.

To investigate the upper bound of the proposed method, we conduct an ideal experimental setting, where there is no negative sample existing in the neighborhood. Specifically, we explicitly involve the ground truth labels of all images in the testing phase, and the samples belong to other identities will be discarded from the neighborhood. After manually removing the negative samples, the results on Occluded-DukeMTMC are improved by 3.7\% mAP and 5.9\% Rank-1 accuracy, respectively. 

\noindent \textbf{Results on Occluded-REID \& Partial-REID} Each person in the two datasets only has 5 gallery images, which is much smaller than Occluded-DukeMTMC. The small number of gallery images is challenging to our method because there is less identity information of the target person can be obtained within the neighborhood. As shown in Tab.\ref{tab:small_dataset}, our method  achieved highest mAP but lower Rank-1 accuracy than the state-of-the-art approaches on these two small datasets. Compared with others, we obtain 2.4\% and 4.5\% mAP gains, indicating our method can find all images of a person more easily. Because of the mistakenly involving negative samples, the results of CMC are lower than some methods (\textit{e.g.}, DSR \cite{DSR} and PVPM \cite{Gao_2020_CVPR}).

\noindent \textbf{Results on Market1501 \& DukeMTMC}
We also apply the method to the traditional ReID datasets (\textit{e.g}, Market1501 and DukeMTMC) in which the person images are assumed to be non-occluded. Specifically, the neighbours of a non-occluded image are searched by the sub-features of upper body because they are more discriminative. As shown in Tab.\ref{tab:holistic}, the proposed method achieves competitive results on the two datasets, especially on DukeMTMC. Actually, under the conditions of non-occlusion, the neighbours of a person image are exploited to enhance the original representation rather than reconstruct it.

\begin{table}[t]
 \begin{center}
  \begin{tabular}
    {l|c |c| c| c}
    \hline
    \multicolumn{1}{l|}{\multirow{2}{*}{\textbf{Method}}}&\multicolumn{2}{c|}{\textbf{Market1501}}&\multicolumn{2}{c}{\textbf{DukeMTMC}}  \\   
    \cline{2-5}
    &mAP&Rank-1&mAP&Rank-1 \\ 
    \hline
    \hline
    PCB&77.3&92.4&65.3&81.9 \\
    Deep CRF&81.6&93.5&69.5&84.9 \\
    MGN&86.9&\textbf{95.7}&78.4&88.7 \\
    BagOfTricks&85.9&94.5&76.4&86.4 \\
    OSNet&84.9&94.8&73.5&88.6 \\
    Pyramid&88.2&\textbf{95.7}&79.0&89.0 \\
    ABD-Net&88.3&95.6&78.6&89.0 \\
    PLR-OSNet&\textbf{88.9}&\textbf{95.7}&81.2&\textbf{91.6} \\
    \hline
    \hline
    Baseline &85.9&94.1&77.3&87.4 \\
    Ours&86.5&94.5&\textbf{83.1}&88.3 \\
    \hline
 \end{tabular}
 \caption{Results on non-occluded datasets of Market1501 and DukeMTMC. We compare wth PCB \protect\cite{Sun_2018_ECCV}, Deep CRF \protect\cite{DeepCRF}, MGN \protect\cite{Wang2018LearningDF}, BagOfTricks \protect\cite{prev_5}, OSNet \protect\cite{OSNet}, Pyramid \protect\cite{zheng2019pyramidal}, ABD-Net \protect\cite{ABD}, PLR-OSNet \protect\cite{PLROSnet}.}
 \label{tab:holistic}
 \end{center} \vspace{-0.5em}
\end{table}

\begin{table}[t]
 \begin{center}
  \begin{tabular}
    {c||l|c |c}
    \hline
    \textbf{No.}&\textbf{Method}&mAP&Rank-1 \\
    \hline
    \hline
    \multirow{2}{*}{0}&Baseline&42.3&48.7 \\
    &Ours&64.2&67.6 \\
    
    \hline
    \hline
    
    \multirow{1}{*}{1}&Base.+OAN&44.0&59.1 \\
    \multirow{1}{*}{2}&Base.+GNN&52.3&56.8 \\
    \multirow{1}{*}{3}&Base.+OAN+Avg-Agg.&61.2&64.3 \\

    \multirow{1}{*}{4}&Base.+OAN+GNN&62.9&66.1\\

    \multirow{1}{*}{5}&Base.+OAN+OR-GNN&64.2&67.6 \\
    \hline
    \hline
    \multirow{3}{*}{6}
    &Base.+RR&58.8&59.1 \\
    &Base.+OAN+RR&63.4&67.5 \\
    &Ours+RR&67.3&68.9 \\
    \hline
 \end{tabular}
 \caption{Ablation studies on Occluded-DukeMTMC.}
 \label{tab:ablation_study} 
 \end{center}\vspace{-1em}
\end{table}

\subsection{Ablation Study}

\noindent \textbf{Effectiveness of Occlusion-Aware Network (OAN)} 
To demonstrate the importance of estimating the occlusion part of a person's image, we use all sub-features to search neighborhoods without using the occlusion state predicted by OAN. The results are shown in Tab.\ref{tab:ablation_study}-2. Compared with Tab.\ref{tab:ablation_study}-4, the results decrease dramatically without using OAN. In particular, the mAP and Rank-1 accuracy decrease from 62.9\% and 66.1\% to 52.3\% and 56.8\%, respectively. Besides, using the non-occluded feature representation is more accurate for image retrieving than using all the part features. We only employ the non-occluded features to retrieve images without representation reconstruction stage, which can significantly improve the baseline Rank-1 accuracy as shown in Tab.\ref{tab:ablation_study}-1.

\noindent \textbf{Effectiveness of Outlier-Aware Module} As shown in Tab.\ref{tab:ablation_study}-4,5, we quantitatively evaluate the effectiveness of the outlier-aware module. Without the help of the outlier-aware module, the mAP and Rank-1 accuracy is decreased by 1.3\% and 1.5\%, respectively, indicating it can alleviate the impact of outliers involved in the neighborhood.

\noindent \textbf{Effectiveness of Feature Reconstruction}
Feature reconstruction is the most important stage in the proposed method, which can bridge the gap between occluded and non-occluded images to make the features comparable. As shown in Tab.\ref{tab:ablation_study}-1,4, the performance significantly improves by reconstructing the representation by images' neighborhood. In greater detail, the mAP and Rank-1 accuracy is increased by 18.9\% and 7.0\%, respectively. To evaluate the effectiveness of the feature reconstruction, we also conduct two variants. One is about using GNN for feature reconstruction, and the other is to average the features of the neighborhood to obtain the reconstructed representation directly. As shown in Tab \ref{tab:ablation_study}-3,4 using GNN outperforms the method of average aggregation, indicating the necessity of the parameter learning in feature reconstruction. 



\begin{figure}[tbp]
    \centering
    \includegraphics[width=1.0\linewidth]{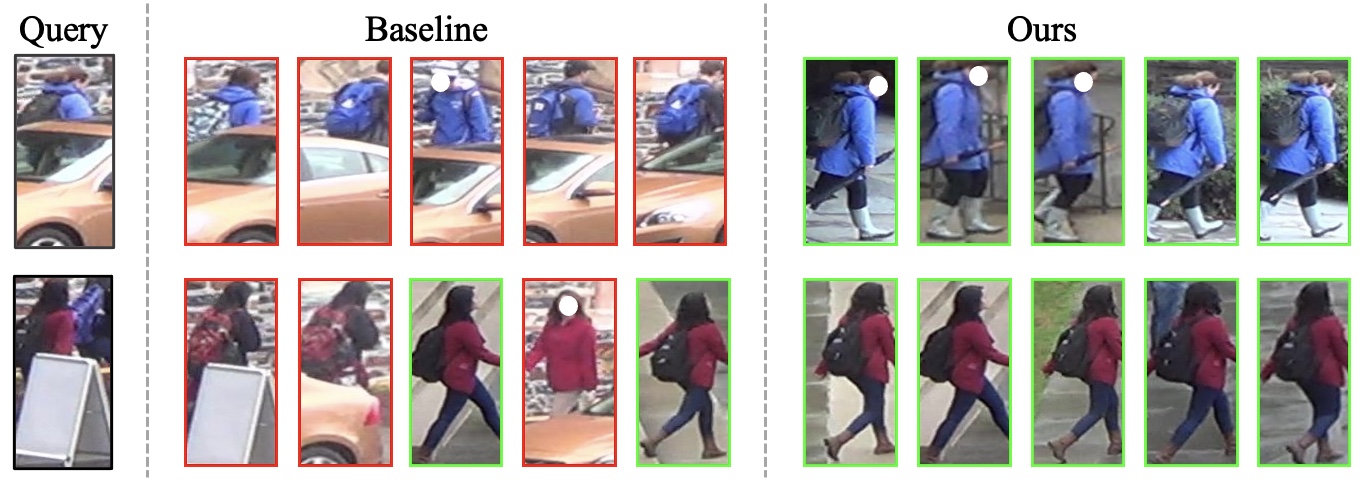}
    \caption{Examples of retrieval results by Baseline and our method. The images with green box denote the positive samples while those with red box denote the negative samples.}
    \label{fig:visualization}\vspace{-0.5em}
\end{figure}

\subsection{Model Analysis}

\noindent \textbf{ReRanking v.s. Ours} It seems that our proposed method using feature reconstruction is related to the reranking method in \cite{Zhong_2017_CVPR} because both of them exploit neighborhood information to improve the ranking performance. The re-ranking cannot handle the occlusion very well as reported in Tab.\ref{tab:ablation_study}-6, where the results of Base.+RR is worse than Ours by 5.4\% mAP and 8.5\% Rank-1 accuracy. Furthermore, our method and re-ranking are complementary, ours+RR still outperforms Ours by 3.1\% mAP and 1.3\% rank-1 accuracy.

\noindent \textbf{Visualization}
As illustrated in Fig. \ref{fig:visualization}, the baseline is more sensitive to occlusions, where the retrieval results easily include negative images with similar obstacles. While the results of our method are more accurate, which can even find non-occluded images given an occluded query image. 

\section{Conclusion}
We have presented a new perspective that the representation of an occluded person image can be reconstructed by its neighborhoods to tackle the problem of occluded person Re-ID. We designed the  occlusion-aware part-based representation, and performed feature reconstruction by the outlier-
removable graph neural network. Experimental results verified the effectiveness of our method on occluded person Re-ID.

\bibliographystyle{named}
\bibliography{ijcai21}
\end{document}